\documentclass{article}
\usepackage{iclr2027_conference}
\usepackage{lmodern}

\usepackage{amsmath,amssymb,amsthm,booktabs,graphicx,microtype,url}
\usepackage[hidelinks]{hyperref}
\usepackage{multirow,subcaption,algorithm,algorithmic}
\usepackage[section]{placeins}

\theoremstyle{plain}
\newtheorem{theorem}{Theorem}

\newtheorem{proposition}[theorem]{Proposition}
\newtheorem{corollary}[theorem]{Corollary}
\theoremstyle{definition}
\newtheorem{definition}[theorem]{Definition}

\newcommand{\E}{\mathbb{E}}

\newcommand{\MDL}{\mathrm{MDL}}

\title{Robot World Models Are Not Invariant\\
to How the Actions Are Written}
\newcommand\affilfootnote[1]{%
  \begingroup\renewcommand\thefootnote{}\footnote{#1}\addtocounter{footnote}{-1}\endgroup}

\iclrfinalcopy

\author{\textbf{Ahmed Karim}$^{1,3}$ \qquad \textbf{Leon Chlon}$^{1,2}$ \\[4pt]
{\normalfont\small $^{1}$Hassana Labs \quad $^{2}$University of Oxford \quad
$^{3}$University College London}}

\begin{document}
\maketitle
\lhead{Preprint}
\affilfootnote{Correspondence to: Leon Chlon $<$\texttt{lc574@cantab.ac.uk}$>$.}

\begin{abstract}
A robot policy is trained with one of two action parameterizations: absolute joint targets, or
deltas relative to the current state. The choice is a live engineering decision in robot
learning, and a world model conditioned on actions inherits it silently. We show that the
inheritance is catastrophic. A latent dynamics model trained on one parameterization and
handed the \emph{identical commanded trajectory} written in the other collapses: retrieval
degrades by $2.6$--$13.4\times$ across three robot datasets and two morphologies,
goal-conditioned action selection falls from $53\%$ to $15\%$, and on PushT the two beliefs
about the same future are near-orthogonal ($\cos=0.067$, worst case $-0.377$), so the
predictor does not degrade gracefully, it answers a different question. This is not a
distribution-shift artifact in the usual sense: the two encodings are mutually reconstructible
at $R^2=0.996$ given the joint input, so no information is lost in the change, and we give the
test that establishes this and separates a valid re-parameterization from a lossy summary or a
sensor swap. The test rejected three of the four axes we proposed. The defect lives in the
\emph{action} channel, which the invariance literature for visual models does not examine:
work there concerns crops, jitter and camera pose, while the parameterization of the commands
themselves goes unaudited. The repair is averaging over the two encodings, and where it goes
matters. Averaging the objective restores task performance by itself, and we report that
rather than crediting it to anything of ours; averaging the \emph{outputs}, which is safe for
probabilities by concavity, is not available for direction-valued prediction, where we show
the normalized mean can score below every member of the orbit. What objective-averaging leaves
behind is the tail: worst-case agreement stays at $0.78$, a disagreement penalty closes it to
$0.995$, and over a latent rollout it is the difference between a worst case that erodes and
one that holds. On PushT, averaging alone does not repair the axis at all.
\end{abstract}

\section{Introduction}

A robot policy has to commit to how it writes an action. Position control emits absolute joint
targets; delta control emits offsets from the current state. Both are standard, both appear in
widely used datasets and codebases, and for a given task the choice is made on grounds that
have nothing to do with learning dynamics: the controller's interface, the teleoperation rig,
what the previous project used. The two describe the same commanded trajectory, and given the
current state $s_t$ each determines the other exactly, since
$a_{\mathrm{rel}}=a_{\mathrm{abs}}-s_t$.

An action-conditioned world model inherits that choice without ever being told it was made. It
receives $(z_t,a)$ and predicts the future, and nothing in its training signal indicates that
$a$ is a \emph{notation} for a trajectory rather than the trajectory itself. This paper asks
what the inheritance costs, and the answer is that it is close to total. A predictor trained on
one parameterization and handed the identical commanded trajectory in the other collapses:
retrieval degrades by up to $13.4\times$, goal-conditioned action selection falls from $53\%$
to $15\%$, and on one dataset the two beliefs about the same future are near-orthogonal. On its
worst input the predictor's two answers point in opposite directions.

\paragraph{Why this is not the usual brittleness story.} Two things separate it. The
information content is provably unchanged: given the joint input the encodings reconstruct
each other at $R^2=0.996$, so there is no sense in which the second presentation tells the
model less, and degradation is not the correct response to it. And the failure is not
graceful. A model pushed off-distribution normally gets noisier; this one answers a different
question. The distinction matters because it decides whether the behaviour should be repaired
or accepted, and \S\ref{sec:admissible} gives the test that makes it, which also rejected
three of the four candidate axes we proposed before this one.

\paragraph{The channel nobody audits.} Work on invariance in visual models is, almost without
exception, about the pixel side: crops, colour jitter, camera pose, augmentation policies.
That is where robustness is measured and where equivariant architectures are built
\citep{cohen2016,egnn}. The action channel is left alone, and it is the channel the robot is
actually commanded through.

Our setting isolates it. World models differ in how much of the learning signal comes from
pixels: the Dreamer line reconstructs observations as part of its objective
\citep{dreamerv3}, while decoder-free and joint-embedding models do not
\citep{tdmpc2,ijepa,vjepa2,lecun2022}. We are in the second camp, and that is not itself a
contribution, since it describes a good deal of recent work. What it buys is diagnostic. When
a reconstruction term is present, the pixel signal constrains the latent directly and the
action channel is not carrying the conditioning alone, so an action-side defect is partly
masked by supervision from elsewhere. Here the encoder is frozen, never inverted, and never
updated: the only thing the predictor learns is how actions move latents. Anything we measure
is therefore a property of the transition data rather than of a rendering, an augmentation
policy, or an encoder trained alongside the defect.

\paragraph{Contributions.}
\begin{itemize}
\item \textbf{A failure mode in action-conditioned world models} (\S\ref{sec:axis}). Absolute
versus relative action parameterization, a choice made for controller-interface reasons,
determines whether a learned dynamics model can use the commands at all. Three robot datasets,
two morphologies, and on PushT the two beliefs about one trajectory are near-orthogonal rather
than merely degraded.
\item \textbf{A test that separates a re-parameterization from a lossy one}
(\S\ref{sec:admissible}). Predictive parity is necessary, and
Theorem~\ref{thm:parity} prices its failure exactly; mutual reconstructibility is a sufficient
certificate rather than a gate; and both must be evaluated on the joint input, since the
action channel alone reports a confident false negative ($R^2=0.757$ against $0.996$). The
test rejected three of the four axes we proposed, which is what makes the fourth worth
reporting.
\item \textbf{The repair, and an honest account of what each part buys}
(\S\ref{sec:repair}). Averaging the objective over the two encodings restores task performance
on its own. Averaging the \emph{outputs} does not transfer from the probability case: we show
by construction that the normalized orbit mean of direction-valued predictions can score below
every member of the orbit (Proposition~\ref{prop:no-jensen}). A disagreement penalty adds no
average accuracy over objective-averaging, and we say so; what it supplies is the tail,
$0.78\to0.995$ worst case, and a worst case that holds rather than erodes over a rollout.
\item \textbf{Evidence in the transition data, not the pixels} (\S\ref{sec:exp}). The encoder
is frozen and never inverted, so nothing here is an augmentation or rendering effect: the
defect is in the states and actions.
\end{itemize}

\section{Why a predictor has no reason to respect a notation}
\label{sec:framework}

Nothing in the training objective distinguishes a notation from a fact. A model fit by log
loss on $(z_t, a_{\mathrm{abs}})$ pairs will use whatever in $a_{\mathrm{abs}}$ predicts the
future, including the part that is an artifact of writing commands as absolute targets rather
than deltas. If that artifact correlates with the answer on the training distribution, using
it shortens the description, and the optimizer does what it was asked to do.

That intuition has an exact price. Write $Z$ for what the two encodings share and $G$ for the
coordinate that distinguishes them, and let $I^{\mathcal F}_{\hat P}(Y;G\mid Z)$ be the
per-sample code length a predictor class gives up by being restricted to predictors that
cannot see $G$. Under two-part MDL the comparison between the restricted and unrestricted
class is
\begin{equation}
\label{eq:threshold}
\mathrm{MDL}_n(\mathcal F_{\mathrm{inv}})-\mathrm{MDL}_n(\mathcal F)
=\underbrace{L(\mathcal F_{\mathrm{inv}})-L(\mathcal F)}_{\text{architectural}}
+\ n\cdot I^{\mathcal F}_{\hat P}(Y;G\mid Z),
\end{equation}
so ignoring the notation is preferred only while the architectural saving covers the
per-sample term (Appendix~\ref{app:formal-setup}). Two consequences matter here. The
per-sample term is what decides whether more data helps, and it takes three distinct values
depending on the transformation, which is why \S\ref{sec:admissible} has to establish which
case applies before any measurement means anything. And the architectural term is the reason a
representation that already discharges the distinction costs nothing to make invariant: an
encoder that never sees absolute joint targets cannot prefer them. Our axis is the case where
no such representation is available, since both parameterizations are natural and neither is
canonical.

\begin{definition}[Make/break criterion]
\label{def:makebreak}
Let $M$ be a task metric and $A$ an orbit-dispersion measure. Compute bootstrap confidence
intervals for $\Delta M$ and $\Delta A$ on held-out data. Report \textsc{make} if the
interval for $\Delta A$ is strictly negative and the interval for $\Delta M$ is bounded
below by $-\varepsilon$; otherwise \textsc{break}.
\end{definition}

\noindent Orbit-constancy always reduces dispersion but need not preserve accuracy, and a
drop is the criterion rejecting the proposed invariance.

\section{Is it a re-parameterization, or is it a different input?}
\label{sec:admissible}

Enforcing orbit-constancy over a transformation that relabels the input is a projection onto
invariants; doing it over one that changes \emph{what is known} is blurring. The two are
indistinguishable from the resulting number, so a dispersion measurement is interpretable only
alongside a verdict on the transformation that produced it.

Both properties we test follow from one observation. A candidate symmetry arrives as a family
of \emph{views} $\varphi_1,\dots,\varphi_K$ carrying the situation $X$ into a common
presentation space, the predictor receiving one and not told which. It is orbit-constant
exactly when $f(\varphi_v(x))=f(\varphi_w(x))$ for all $x,v,w$, that is, when it is measurable
with respect to the quotient of that space by the relation those equalities generate. Write
$Q$ for the class of $\varphi_v(X)$ under that relation, which does not depend on $v$, and
$R(\varphi_v):=H(Y\mid\varphi_v(X))$ for the best risk from view $v$ alone.

\begin{theorem}[Parity is necessary, and the parity gap is its price]
\label{thm:parity}
Let $R_{\mathrm{inv}}$ be the best log loss achievable by an orbit-constant predictor. Then
$R_{\mathrm{inv}} = H(Y\mid Q)$ and
\[
R_{\mathrm{inv}}\ \ge\ \max_{v} R(\varphi_v),
\qquad\text{so}\qquad
R_{\mathrm{inv}}-\min_{v}R(\varphi_v)\ \ge\ \delta
:=\max_{v}R(\varphi_v)-\min_{v}R(\varphi_v).
\]
\end{theorem}

\noindent The proof is three lines (Appendix~\ref{app:parity}): $Q$ is a function of
$\varphi_v(X)$ for \emph{every} $v$, so conditioning on it can only raise risk, and
orbit-constancy is therefore at least as costly as the worst view in the family. The
consequences are the test.

\paragraph{Predictive parity (necessary).} Used in isolation the views must be equally
predictive of $y$, and Theorem~\ref{thm:parity} says what a failure costs: the parity gap
$\delta$ is a lower bound on what enforcing orbit-constancy gives up against simply using the
better view. This is why \emph{channel choices} and \emph{lossy summaries} must be rejected
rather than repaired. It also prices the axes we discarded: the two cameras differ by
$6.6$ against $23.3$ mean rank, so enforcing constancy across them buys agreement by
discarding the better sensor.

\paragraph{Reconstructibility (sufficient, not necessary).} If each view is recoverable from
the other then $Q$ generates the same $\sigma$-algebra as each $\varphi_v(X)$, so
$R_{\mathrm{inv}}=R(\varphi_v)$ and orbit-constancy is free: bijective relabelling is costless,
which is the certificate. It is not necessary. A \emph{quotient} symmetry discards a nuisance
coordinate by construction, and from $\theta+2\pi k$ one cannot recover $k$, yet
$\delta=0$ because $k$ says nothing about $y$, so Theorem~\ref{thm:parity} still gives zero
cost. Requiring bijectivity would reject valid symmetries, so we report it as a certificate
rather than a gate.

\paragraph{Evaluate on the joint input.} Admissibility is a property of what the predictor
receives, not of one channel. Absolute versus relative action encoding \emph{fails}
reconstruction on the action channel alone ($R^2=0.757$) and \emph{passes} jointly with the
state ($R^2=0.996$), because $a_{\mathrm{rel}}=a_{\mathrm{abs}}-s_t$ is a bijection only
given $s_t$. Testing a channel in isolation produces a confident false negative.

In our own use the test rejected three transformations we had proposed: photometric jitter
(an augmentation, not a group action), camera choice (different sensors carry different
information: one view reached rank $6.6$ where another reached $23.3$), and
action-window summaries (\texttt{first}/\texttt{mean}/\texttt{mean\_std}/\texttt{concat}
form a lossy hierarchy, task $R^2$ $0.986$ versus $0.709$).
The test also partitions our own experimental settings, six of which are group-structured and
receive a certificate while the four action-conditioning settings are \emph{action ablations}
rather than orbit-constancy tests: action-shuffle permutes actions across examples, generates
no orbit, and so has no dispersion measure for Definition~\ref{def:makebreak} to consult
(Appendix~\ref{app:cross}). That leaves action-conditioned dynamics without an orbit-constancy
test at all, which is where the experiments begin.

\section{Repair: averaging over the two encodings}
\label{sec:repair}

The repair is orbit averaging. Everything before this section exists to establish what one is
entitled to average over, because averaging is a single line and the entitlement is not. What
remains is a choice of \emph{where} the average goes, and the output space decides it.

Given a certified axis, let $\{\phi_v\}$ enumerate the equivalent presentations. We train a
single predictor $g_\theta$ to fit the target under every presentation while agreeing with
itself across them:
\begin{equation}
\label{eq:repair}
\mathcal L(\theta)=\frac{1}{K}\sum_{v}\ell\!\left(g_\theta(z,\phi_v(a)),\,y\right)
\;+\;\lambda\cdot\frac{2}{K(K-1)}\sum_{v<w}
d\!\left(g_\theta(z,\phi_v(a)),\,g_\theta(z,\phi_w(a))\right),
\end{equation}
with $\ell$ the task loss (InfoNCE against the true future for retrieval targets, squared
error for regression) and $d$ a disagreement penalty. The first term is an average over the
orbit, so this \emph{is} orbit averaging; what has moved is its position. The average is taken
over the loss rather than over the predictions, and the predictor is never asked to match a
summary of its own outputs. It is asked to be right under every presentation, and to be
consistent across them.

\paragraph{Where the average belongs.} The natural alternative puts it on the outputs: form
the orbit mean of a realization-sensitive teacher's predictions and fit a student to it. Where
the prediction is a probability that is not merely reasonable but guaranteed, writing
$\bar p:=\frac1K\sum_k f(\cdot\mid g_k\cdot x)$, the empirical Jensen gain
\[
\hat J(x;y):=\frac1K\sum_{k=1}^{K}\big(-\log f(y\mid g_k\cdot x)\big)-\big(-\log\bar p(y)\big)\ \ge\ 0
\]
is non-negative by concavity of the logarithm (Appendix~\ref{app:diagnostics}), so the
mixture's log loss is never worse than the average of its members'. Averaging the outputs is
therefore the right move whenever the prediction is a probability, and the point of what
follows is that our setting is not that case.

The guarantee is a property of the output space, not of the symmetry, and it does not survive
the move to a direction in a learned representation. There the target is a unit vector, the
discrepancy is an angle, and the mean of unit vectors that disagree in direction is short and
points at the centroid, so renormalising it need not return a member of the orbit. The failure
is not a corner case.

\begin{proposition}[No Jensen guarantee for direction-valued prediction]
\label{prop:no-jensen}
Let predictions and target lie on $S^{d-1}$ with score $\cos(\hat u,y)$. There exist a finite
orbit and predictions $\{u_k\}$ for which the normalized mean
$\bar u=\sum_k u_k/\|\sum_k u_k\|$ scores strictly below every $u_k$, and others for which
$\sum_k u_k=0$ and $\bar u$ is undefined.
\end{proposition}

\noindent A witness for the first is $d=2$, $y=(1,0)$ and
$u_1=(1,0)$, $u_{2,3}=(-0.9,\pm\sqrt{1-0.81})$: the members score $1,-0.9,-0.9$ while
$\bar u=(-1,0)$ scores $-1$. For the second, take $u_{1,2}=(0,\pm1)$. The first witness is
stable, since the inequality is strict, so no genericity assumption removes it; the second is
not, but $\bar u$ is ill-conditioned throughout a neighbourhood of it
(Appendix~\ref{app:diagnostics}). This is why output averaging is the right move
wherever the prediction is a probability and fails here: Table~\ref{tab:repair} shows
the orbit-mean student at rank $11.55$ against $5.25$ for Eq.~\eqref{eq:repair}.
Eq.~\eqref{eq:repair} sidesteps the issue by never forming the mean. The predictor is asked to
fit the target under each presentation and to agree with itself, which is well defined
whatever space the prediction lives in.

\section{Experiments}
\label{sec:exp}

\S\ref{sec:admissible} left action-conditioned dynamics without an orbit-constancy test. We
supply one, measure it on three robot datasets across two morphologies, and close with what
decides whether such a gap exists at all.

\subsection{Absolute versus relative action encoding}
\label{sec:axis}

We predict future JEPA latents from current latents and an action window,
$\hat z_{t+\Delta}=g_\theta(z_t,\phi(a_{t:t+\Delta}))$, with a frozen I-JEPA encoder
\citep{ijepa} and no pixel prediction \citep{lecun2022}. The axis is the \emph{encoding of
the action window}:
\[
A:\ a_{t:t+\Delta}\ \text{(absolute joint targets)}
\qquad
B:\ a_{t:t+\Delta}-s_t\ \text{(relative to current state)}.
\]
Both describe the same commanded trajectory, and position control versus delta control is a
live design choice in robot policy learning, so the axis is not contrived. It is admissible:
evaluated on the joint input as \S\ref{sec:admissible} requires, the two encodings are
mutually reconstructible at $R^2=0.996$ with a task-$R^2$ gap of $0.003$. Reconstruction from
the action channel alone reaches only $R^2=0.757$, which is the false negative that section
warns about.

Holding the camera fixed keeps all predictions in one latent space, so dispersion is
well defined. A multi-camera axis would not be, since predictions would live in
camera-specific spaces.
Following Definition~\ref{def:makebreak}, the teacher is \emph{realization-sensitive}:
trained on the canonical encoding alone, then evaluated across the orbit.

\begin{figure}[tbp]
\centering
\includegraphics[width=0.88\textwidth]{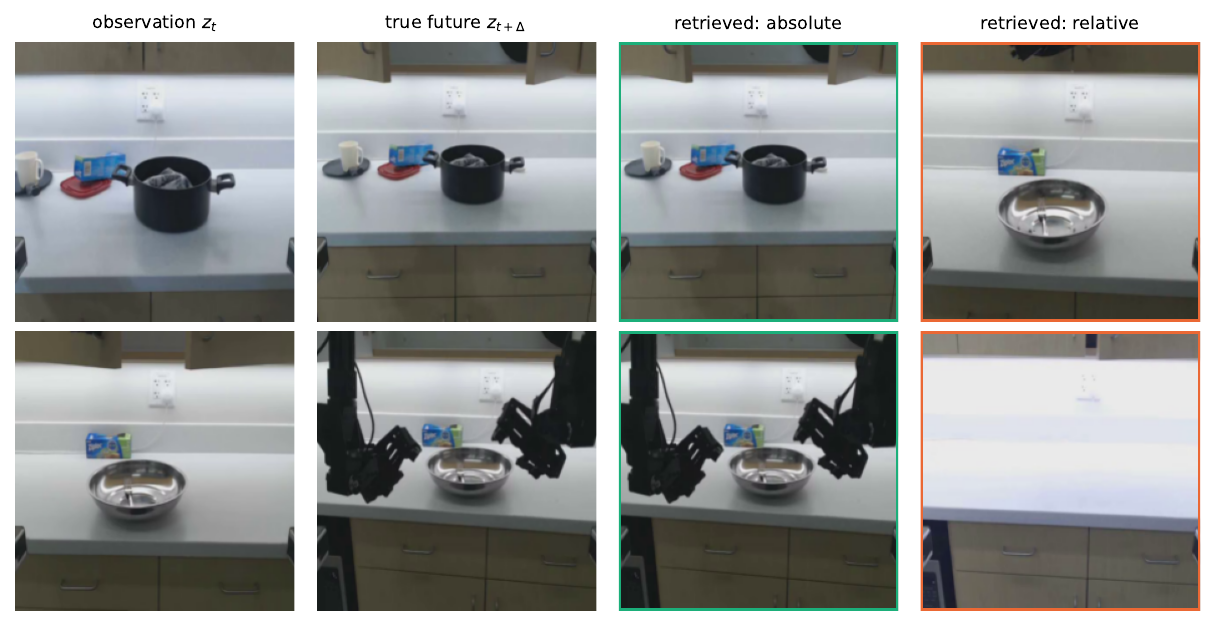}
\caption{The same failure on frames. Each row is one test instance: the observation at $t$,
the true frame at $t+\Delta$, and the frames retrieved when the predictor is handed the
identical action window in absolute and in relative encoding. Green marks a correct
retrieval, orange an incorrect one. Rows are restricted to misses from a \emph{different}
episode, since a same-episode miss is a near-duplicate frame and would show the reader
nothing. Of $384$ test instances, $125$ are absolute-correct and relative-wrong, $95$ of them
from a different episode; two are shown.}
\label{fig:montage}
\end{figure}

Table~\ref{tab:gap} reports the measurement: the identical action window in the equivalent
encoding collapses retrieval by $2.6$--$13.4\times$. The two ALOHA datasets agree to three
decimals in cross-encoding cosine, so the effect is not task-specific, and on PushT, a
different robot with $2$-dimensional actions, the two beliefs about the same future are close
to orthogonal. Figure~\ref{fig:montage} shows it on frames: handed the same trajectory in the
other notation, the predictor retrieves a visibly different scene.

\paragraph{A certified special case of covariate shift.} We should be exact about what this
is. The protocol trains on one encoding and tests on the other, so it is a distribution
change, and we claim no new phenomenon. What the certificate adds is the thing covariate-shift
results normally cannot supply. Such results confound \emph{different distribution} with
\emph{less information}, and the two have opposite implications: if the shifted input carries
less about the target, degradation is correct behaviour and there is nothing to repair.
Admissibility excludes that branch by construction ($R^2=0.996$ jointly, task-$R^2$ matched to
$0.003$), turning a claim about brittleness into a claim about brittleness \emph{to a change
that provably carries no information about the target}. Whether repair is then free in
practice is a separate question, and \S\ref{sec:repair} finds that it is not on the dataset
where the collapse is largest. The magnitude is what makes it worth acting on: on PushT the cosine is $0.067$ and the worst case is
negative, so the predictor does not degrade gracefully, it answers a different question. The
symmetry is one of the data pipeline rather than of the world, with no conservation law behind
it, which is why it had to be certified rather than assumed.

\begin{table}[tbp]
\centering
\caption{Orbit dispersion under the action-encoding axis. $\Delta=4$s, 3 seeds,
episode-disjoint splits, pool size 384. Collapse is held-out rank over canonical rank; $\cos$ is agreement between the two
encodings' predictions, mean and worst case.
ALOHA datasets \citep{aloha}; PushT \citep{pusht}; all via \citet{lerobot}.}
\label{tab:gap}
\begin{tabular}{llrrrrr}
\toprule
& & \multicolumn{2}{c}{Retrieval rank} & & \multicolumn{2}{c}{$\cos$} \\
\cmidrule(lr){3-4}\cmidrule(lr){6-7}
Dataset & Morphology & Canon. & Held out & Collapse & mean & worst \\
\midrule
\texttt{aloha\_mobile\_cabinet} & 14-D bimanual, 50\,fps & 6.19 & 61.5 & $9.9\times$ & 0.370 & $-0.178$ \\
\texttt{aloha\_static\_coffee}  & 14-D bimanual, new task & 3.97 & 53.3 & $13.4\times$ & 0.371 & $-0.096$ \\
\texttt{pusht}                  & 2-D planar pusher, 10\,fps & 66.2 & 169.8 & $2.6\times$ & 0.067 & $\mathbf{-0.377}$ \\
\bottomrule
\end{tabular}
\end{table}

\paragraph{The failure is a control failure.} Retrieval rank measures representation quality;
control asks which action reaches a goal, scoring candidates under the learned dynamics in the
same encoding as the query. The teacher selects the correct action $53.1\%$ of the time in its
own encoding and $15.4\%$ in the equivalent one (Table~\ref{tab:repair}), and the mean cosine
of $0.370$ understates the tail, whose worst case is $-0.178$. A model that cannot choose an
action when the same commands are written differently is not usable for control, whatever its
retrieval score.

\begin{figure}[tbp]
\centering
\includegraphics[width=\textwidth]{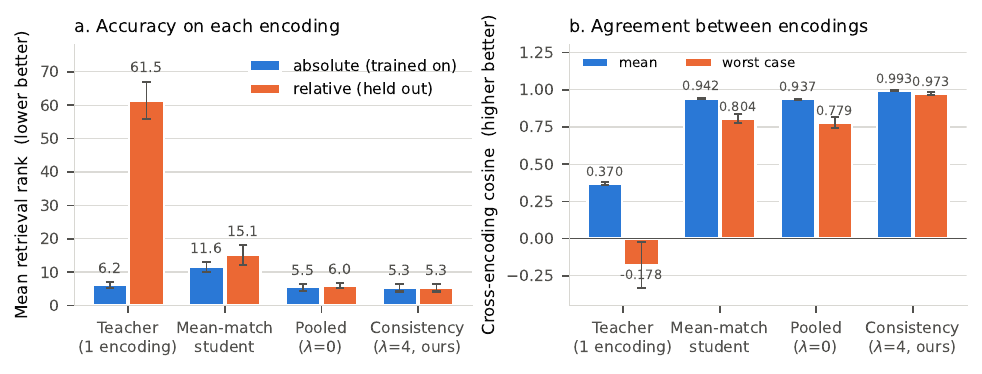}
\caption{Left: retrieval rank on each encoding. Right: cross-encoding agreement, mean and
worst case over the evaluation set. The teacher is competent on the encoding it was trained on
and collapses on the equivalent one. Pooled training ($\lambda=0$) is shown because it, not
mean-matching, is the baseline that competes: it recovers accuracy but leaves worst-case
agreement at $0.779$, which only the consistency term lifts. The teacher's worst case is
$-0.178$: on its worst input the two notations point in opposite directions.}
\label{fig:main}
\end{figure}

\begin{table}[tbp]
\centering
\caption{Repair on \texttt{aloha\_mobile\_cabinet}, 3 seeds, all rows from the same runs.
$\lambda$ is the disagreement weight of Eq.~\eqref{eq:repair}; $\lambda=0$ is the same
two-view pooled training with the penalty off, which is the baseline any practitioner would
reach for once the axis is known; the orbit-mean row fits a student to the averaged teacher
prediction instead. $\cos$ is mean and worst-case pairwise agreement across the orbit. Action
selection is top-1 over 32 candidates, chance $0.031$.}
\label{tab:repair}
\begin{tabular}{lrrcccc}
\toprule
 & \multicolumn{2}{c}{Retrieval rank} & \multicolumn{2}{c}{Agreement} & \multicolumn{2}{c}{Action selection} \\
\cmidrule(lr){2-3}\cmidrule(lr){4-5}\cmidrule(lr){6-7}
 & Canon. & Held out & mean & worst & Canon. & Held out \\
\midrule
Teacher            & 6.19 & 61.5 & 0.370 & --- & $0.531$ & $0.154$ \\
Orbit-mean student & 11.55 & 15.1 & 0.942 & 0.804 & --- & --- \\
$\lambda=0$        & 5.46 & 5.98 & 0.937 & 0.779 & $\mathbf{0.443}$ & $0.401$ \\
$\lambda=1$        & 5.51 & 5.58 & 0.981 & 0.939 & $0.408$ & $0.406$ \\
$\lambda=4$        & 5.25 & 5.31 & 0.993 & 0.973 & $0.426$ & $0.402$ \\
$\lambda=16$       & \textbf{5.17} & \textbf{5.19} & \textbf{0.999} & \textbf{0.995} & $0.411$ & $\mathbf{0.411}$ \\
\bottomrule
\end{tabular}
\end{table}

\paragraph{Repair, and what each part of it buys.} Table~\ref{tab:repair} separates three
things that are easy to conflate, and the separation is not flattering to the penalty, so we
state it first. \emph{Averaging the objective over the orbit recovers the task on its own.}
Training on both encodings with the penalty off ($\lambda=0$) takes held-out rank from $61.5$
to $5.98$ and held-out action selection from $15.4\%$ to $40.1\%$; $\lambda=16$ moves those to
$5.19$ and $41.1\%$, one point against a seed standard deviation of four. On this evidence
\textbf{the penalty does not improve average task performance over averaging alone}, and we
do not claim that it does.

What it buys is invariance in the sense the word is usually meant. At $\lambda=0$ the
worst-case agreement across the orbit is $0.779$: on average the two notations agree, and on
the inputs where they do not the predictions are still far apart. The penalty closes that,
$0.779\to0.995$, a four-fold reduction in worst-case disagreement and well outside seed
noise. The price is about three points of canonical action selection, $44.3\%\to41.1\%$.
So the honest summary is that once the axis is known, augmentation restores accuracy and the
penalty converts ``usually agrees'' into a guarantee, which is what Theorem~\ref{thm:parity}
is stated in terms of and what a certificate needs. Fitting the orbit mean does neither: it
reaches $0.942$ mean agreement at roughly twice the rank error, for the reason
Proposition~\ref{prop:no-jensen} gives.

\paragraph{The mean is stable over a horizon; the worst case is not.} Every number above is
one step, and a world model is used over many. We roll the predictor forward on held-out
episodes for hops of $4$s each, feeding its own output back in with the true action window and
state supplied at every hop, so the only iterated quantity is the model's belief
(Table~\ref{tab:rollout}). This is not closed-loop control, and predictive quality decays over
the horizon (cosine to the true latent $0.33\to0.13$ on cabinet), so later hops describe
agreement about an increasingly uncertain quantity.

On the two ALOHA datasets mean agreement is close to flat for every model, so the disagreement
does not compound in the average and we make no such claim. Worst-case agreement behaves
differently: the teacher erodes, pooled training slows that without stopping it, and only the
consistency term holds. PushT separates the two far more sharply, because there \emph{pooled
training does not repair the axis at all}: mean agreement reaches $0.633$ and worst case is
\emph{negative}, $-0.348$ by the second hop, so a predictor trained on both encodings still
produces opposed beliefs about the same commanded trajectory. The penalty lifts those to
$0.946$ and $+0.139$ without closing the gap. The easier the case, the more averaging alone
suffices; the harder it is, the more of the work the consistency term does.

\begin{table}[tbp]
\centering
\caption{Latent rollout on held-out episodes, 3 seeds, three datasets and two morphologies:
cross-encoding cosine at the final hop (three hops of $4$s, two for PushT, whose episodes are
short). Mean and worst case over the evaluation set. Per-hop trajectories are in
Table~\ref{tab:rollout-full}. Chains need $\text{horizon}\times\Delta$ of runway and so start
earlier in an episode than the pairs of Table~\ref{tab:repair}, so read the comparison across
models, not against that table.}
\label{tab:rollout}
\small
\begin{tabular}{lcccccc}
\toprule
& \multicolumn{2}{c}{Teacher} & \multicolumn{2}{c}{$\lambda=0$} & \multicolumn{2}{c}{$\lambda=16$} \\
\cmidrule(lr){2-3}\cmidrule(lr){4-5}\cmidrule(lr){6-7}
Dataset & mean & worst & mean & worst & mean & worst \\
\midrule
ALOHA cabinet & 0.805 & 0.242 & 0.988 & 0.896 & 1.000 & \textbf{0.997} \\
ALOHA coffee  & 0.898 & 0.614 & 0.991 & 0.955 & 1.000 & \textbf{0.998} \\
PushT         & 0.178 & $-0.497$ & 0.677 & $-0.348$ & \textbf{0.872} & \textbf{0.139} \\
\bottomrule
\end{tabular}
\end{table}

\paragraph{On the value of finding the axis.} That averaging over the orbit works, once you
know which orbit, is not a weakness of the diagnosis; it is the point of it. The averaging is
one term in Eq.~\eqref{eq:repair}. What it presupposes is an admissible transformation to
average over, and three of the four we proposed were not one: had we averaged over them, the
cost would have been the parity gap of Theorem~\ref{thm:parity}, paid for nothing. The
entitlement is the hard part, which is why it occupies \S\ref{sec:admissible} and this
section is short.

\paragraph{Where repair is not free.} On PushT the repaired predictor improves worst-case
rank $169.8\to85.7$ and $\cos 0.067\to0.971$, but costs canonical accuracy
($66.2\to79.2$). That teacher is weak in absolute terms (rank $66$ of $384$), and
\S\ref{sec:threshold} shows the repair is free only where the model has slack.

\subsection{The defect is flat in $n$ over the range we can sweep}
\label{sec:threshold}

The action-encoding axis is regime (iii) of \S\ref{sec:framework}: the teacher is trained on
the canonical encoding only, so the coordinate distinguishing the two notations is never
exercised by a label. Equation~\eqref{eq:threshold} therefore prices nothing along it, and the
prediction to test is that the departure does not close as the model gets better at its own
task. Sweeping $n$ by slicing a fixed cached training set, so that latents are identical
across conditions, the collapse ratio grows monotonically:
$2.5\times,3.9\times,5.4\times,9.9\times$ for $n=150,300,600,1200$
(Figure~\ref{fig:scaling}).

The ratio alone would overstate this, and we read it carefully. Cross-encoding cosine is
\emph{flat} in $n$ ($0.357,0.337,0.374,0.370$): more data does not make the predictor
disagree with itself more. It makes the same disagreement more costly, because the canonical
model sharpens while the held-out one does not, and the ratio is a quotient of the two. The
claim the sweep supports is therefore the weaker and more useful one, that the defect is
invariant to scale rather than amplified by it: at $n=1200$ the predictor is eight times
better on the encoding it saw and no better at all on the one it did not. The same pattern
holds for optimization and for capacity (Table~\ref{tab:axes}), with the same caveat and one
more. The larger and longer-trained teachers are also worse on their own encoding, so both
are partly overfitting $1200$ examples; the cosine comparison is the defensible half, and a
clean capacity claim would require matched canonical accuracy, which we do not have. We do
not read the capacity column as independent evidence.

\paragraph{Locating $I(Y;G\mid Z)$.} Sweeping $\lambda$ (Figure~\ref{fig:lambda}) exposes the
cost term directly. At $n=150$ a strong penalty ($\lambda=256$) is worth $-6.59$ rank; at
$n=1200$ it costs $+0.38$. The sign flips as $n$ grows, as the right-hand side of
Eq.~\eqref{eq:threshold} predicts. Below the crossover orbit-constancy is free
regularization; above it, it is a real cost. The useful operating point,
$\lambda\in[4,16]$, lies below.

\subsection{Why this axis is the hard case}
\label{sec:exp-others}

Whether a re-parameterization costs anything is decided by the representation, not by the
transformation. Two control axes with known analytic structure
(Appendix~\ref{app:certificates}) show dispersion at $10^{-15}$ and $10^{-12}$, which is to
say none, and the reason is not robustness: the Kepler featurization encodes an angle as
$(\sin M,\cos M)$, so $M$ and $M+2\pi k$ are literally the same input, and the Lorenz
predictor never receives absolute time. The transformation never reaches the model. A
molecular-dynamics ablation makes the same point with the variable isolated
(Appendix~\ref{app:molecular}): same system, same data, same criterion, and the verdict flips
from \textsc{make} to \textsc{break} purely by exchanging an invariant encoder for raw
coordinates.

This is the architectural term $L(\mathcal F)-L(\mathcal F_{\mathrm{inv}})$ of
Eq.~\eqref{eq:threshold} made visible: \textbf{a gap exists exactly when the representation
does not already discharge the distinction.} It is also why absolute versus relative action
encoding is the hard case rather than a curiosity. For a camera angle or a time origin one can
featurize the nuisance away, and should. Here both parameterizations are natural, neither is
canonical, and a predictor that takes an action window as a vector of numbers has no way to
know which convention produced it. There is no featurization to reach for, which leaves
measuring the gap and training it away.

\section{Discussion}

\paragraph{Discovering rather than assuming invariance.} The question of which
transformations a predictor should be invariant to is the one invariant risk minimization
\citep{irm} and the causal-invariance line \citep{peters2016} also ask, and the contrast is
informative. Those methods are handed a partition into environments and search
for a representation whose optimal predictor is stable across it, so invariance is
\emph{inferred} from variation the data already contains. We are handed a candidate
transformation and asked whether enforcing constancy over it is projection or blurring, which
is decidable from a single environment and is a property of the transformation rather than of
an environment family. The two are complementary, and neither subsumes the other: environments
say an invariance exists without naming the group, certification says whether a named group is
one. Proposing transformations automatically and certifying them with
Theorem~\ref{thm:parity} is the obvious thing to build next.

\paragraph{Relation to prior work.} Building invariance into the hypothesis class
\citep{cohen2016,egnn} is the left-hand side of Eq.~\eqref{eq:threshold} and is preferable
wherever it applies, since the price is zero; our concern is what it does not cover. The
generalization benefits are established \citep{elesedy2021} and augmentation reads as
approximate marginalization over a nuisance group \citep{chen2020}, which
Eq.~\eqref{eq:threshold} prices. Latent-space world models \citep{ijepa,vjepa2,lecun2022} are
usually evaluated with representation-space metrics and an action-channel ablation, and
comparison with reinforcement-learned ones \citep{dreamerv3,tdmpc2} is orthogonal: we claim
not better forecasting but that a forecast should not depend on which of two equivalent
notations was handed over. We release Mezzanine, which implements the procedure end to end.

\paragraph{Where repair is unavailable.} The repair needs a certified symmetry whose orbit can
be enumerated, and both conditions bind. Where an orbit is large or continuous, enumerating it is
out of reach and no post-hoc repair is available at any price
(Appendix~\ref{app:impossible}). Our own
measurements bracket it: flattened LJ coordinates make the orbit $S_n\times SE(3)$ and
enumeration cannot preserve accuracy, while the action-encoding axis is tractable precisely
because it has two elements.

\paragraph{Limitations.} The predictor is a small MLP on frozen latents and three seeds is
thin, so we report directions rather than magnitudes, and the capacity ablation's larger
teacher is also worse on its own encoding, so it is not independent evidence. Evaluation is
offline: action selection is a proxy for control, and the rollout is closed in the latent only.
The admissibility probe is linear ridge regression, so it certifies linear relabellings
trivially and may reject nonlinear bijections.

\paragraph{Conclusion.} Averaging over the orbit is the repair, and it is the easy part.
Which orbit, whether the transformation is a symmetry at all, and where the average belongs
once the prediction is a direction: those come first, and they are the paper.

\subsection*{Reproducibility statement}

All results regenerate from the accompanying repository. Theorem~\ref{thm:parity} is proved in
Appendix~\ref{app:parity} and Proposition~\ref{prop:no-jensen} in
Appendix~\ref{app:diagnostics}; Appendix~\ref{app:experimental} gives datasets, architecture,
hyperparameters, seeds and hardware. The robotics experiments use public
LeRobot datasets; the exact commands, seeds and run directories are listed in the
repository, latents are cached by world and encoder fingerprint, and every reported number
comes from a stored \texttt{results.json}. Run-to-run variation is reported as standard
deviation over three seeds throughout. We note that ranks vary by 2--4\% between CPU and GPU
from floating-point ordering while the dispersion measures agree to three decimals, so the
compute device is recorded alongside the seed.

\subsection*{Ethics statement}

This work studies failure modes of predictive models and proposes a diagnostic and a repair.
It uses public robot-learning datasets and involves no human subjects, no personally
identifying data, and no new data collection. The diagnostic is intended to be run before a
learned dynamics model is trusted for control rather than after, and we note that the repair
buys worst-case agreement at a measurable cost in average accuracy, so the operating point is
application-specific rather than universal.

\appendix

\section{Experimental detail}
\label{app:experimental}

\paragraph{Datasets.} All three are public LeRobot releases \citep{lerobot}, used unmodified.

\begin{center}\small
\begin{tabular}{llrrrrr}
\toprule
Dataset & Camera & Episodes & Frames & fps & $\Delta$ (steps) & train/test eps. \\
\midrule
\texttt{aloha\_mobile\_cabinet} & \texttt{cam\_high} & 85 & 127{,}500 & 50 & 200 & 68 / 17 \\
\texttt{aloha\_static\_coffee}  & \texttt{cam\_high} & 50 & 55{,}000  & 50 & 200 & 40 / 10 \\
\texttt{pusht}                  & \texttt{image}     & 206 & 25{,}650 & 10 & 40  & 165 / 41 \\
\bottomrule
\end{tabular}
\end{center}

\noindent Splits are \emph{episode-disjoint}: no frame from a test episode appears in
training, so a nearest-neighbour retrieval cannot succeed by recalling an adjacent frame of a
seen trajectory. $\Delta$ is $4$ seconds in every case, which is $200$ steps at $50$\,fps and
$40$ at $10$\,fps.

\paragraph{Encoder.} Frozen I-JEPA ViT-H/14 \citep{ijepa}
(\texttt{facebook/ijepa\_vith14\_1k}), fp16, batch size $32$. Frames are resized to $224$ at
cache-build time. Embeddings are the concatenated mean and standard deviation over patch
tokens of the fourth-from-last block. The encoder is never fine-tuned, never inverted, and no
gradient reaches it; latents are cached and keyed by world and encoder fingerprint together
with the resize, so a run cannot silently mix provenance.

\paragraph{Predictor.} An MLP on $[z_t \,\|\, \phi_v(a_{t:t+\Delta})]$ predicting the
$\ell_2$-normalised future latent: hidden width $1024$, depth $2$, trained $800$ steps with
InfoNCE against the true future, in-batch negatives. The action window is $16$ points sampled
uniformly across the $\Delta$ horizon, so the action input is $16 \times d_a$ ($d_a = 14$ for
ALOHA, $2$ for PushT). Latents are centred using statistics from the training split only.
$n_{\mathrm{train}} = 1200$, $n_{\mathrm{test}} = 384$.

\paragraph{Repair.} Eq.~\eqref{eq:repair} with $\lambda \in \{0,1,4,16\}$, the same $800$
steps, $d$ the mean pairwise cosine disagreement. $\lambda = 0$ is identical two-view pooled
training with the penalty switched off and is the baseline throughout.

\paragraph{Metrics.} Retrieval rank is the position of the true future latent among the $384$
test latents under cosine similarity. Cross-encoding agreement is the cosine between the two
encodings' predictions for the same input, reported as a mean and as the minimum over the
evaluation set. Goal-conditioned action selection scores $32$ candidate action windows under
the learned dynamics against the true future and reports top-1; chance is $0.031$.

\paragraph{Seeds and hardware.} Every number is the mean over three seeds ($0,1,2$) with the
standard deviation where reported; the seed controls split sampling, initialisation and
batching. All runs are on a single RTX 5090. With latents cached, a full
three-seed sweep on \texttt{aloha\_mobile\_cabinet} takes a few minutes; the first run on a
dataset must decode video and takes longer.

\paragraph{What is not controlled.} Three seeds is thin, and we report the direction of each
effect rather than its magnitude. Capacity and optimisation sweeps vary one axis against a
fixed $1200$-example training set, so the larger and longer-trained teachers are partly
overfitting; that is stated where those numbers appear and we do not treat them as independent
evidence.

\section{Action ablations are not orbit-constancy tests}
\label{app:cross}

Action conditioning is routinely validated by \emph{action-shuffle}: permute actions across
examples and check that accuracy collapses. Four of our settings were originally run that way
(ALOHA, LIBERO-10, PushT, iPhyRE), and the test of \S\ref{sec:admissible} says plainly what
that evidence is and is not.

Action-shuffle is deliberately not evidence-preserving. It feeds the predictor an action
window belonging to a different example, so the intervention destroys information rather than
re-presenting it, generates no orbit, and leaves no dispersion measure $A$ for
Definition~\ref{def:makebreak} to consult. Such a result is still informative: a predictor
whose accuracy collapses under shuffled actions is demonstrably using the action channel
rather than ignoring it, and we report those rows as \emph{action ablations}
(Appendix~\ref{app:robotics}). What they are not is evidence that the predictor's belief is
stable under a change that preserves the commands, which is a different and stronger property
and the one this paper measures.

That gap is the reason \S\ref{sec:axis} exists. Before it, action-conditioned dynamics had no
orbit-constancy test at all, only ablations that confirm the channel is read.

Table~\ref{tab:cross} adds four numerical axes with known analytic structure, which serve as
controls: two have a real departure that repair removes, and two have none to begin with.

\paragraph{Repair beyond robotics.} On node permutation, a predictor trained with
Eq.~\eqref{eq:repair} at $\lambda=0$ reaches dispersion $21.2$ against $54.1$ for one fitted
to the orbit mean, with marginally lower test error. On periodic integration it makes no
difference: that predictor already uses an invariant featurization and sits at $10^{-13}$.

\section{Rollout divergence, per hop}
\label{app:rollout}

\begin{table}[htbp]
\centering
\caption{Per-hop cross-encoding cosine for Table~\ref{tab:rollout}: 3 seeds, three datasets and two morphologies;
cross-encoding cosine between the absolute and relative rollouts at each hop. PushT episodes
are short and admit two hops rather than three. Read the trend across hops, not the absolute
values: chains need $\text{horizon}\times\Delta$ of runway and so start earlier in an episode
than the pairs of Table~\ref{tab:repair}, where the encodings are more similar at the input
($\cos 0.46$ against $0.27$).}
\label{tab:rollout-full}
\small
\begin{tabular}{llccc@{\hskip 1.4em}ccc}
\toprule
& & \multicolumn{3}{c}{Mean $\cos$} & \multicolumn{3}{c}{Worst-case $\cos$} \\
\cmidrule(lr){3-5}\cmidrule(lr){6-8}
Dataset & Model & hop 1 & hop 2 & hop 3 & hop 1 & hop 2 & hop 3 \\
\midrule
ALOHA & Teacher      & 0.836 & 0.803 & 0.805 & 0.515 & 0.350 & 0.242 \\
cabinet & $\lambda=0$  & 0.992 & 0.990 & 0.988 & 0.972 & 0.951 & 0.896 \\
 & $\lambda=16$ & 1.000 & 1.000 & 1.000 & \textbf{1.000} & \textbf{0.999} & \textbf{0.997} \\
\midrule
ALOHA & Teacher      & 0.916 & 0.889 & 0.898 & 0.707 & 0.676 & 0.614 \\
coffee & $\lambda=0$  & 0.995 & 0.993 & 0.991 & 0.987 & 0.973 & 0.955 \\
 & $\lambda=16$ & 1.000 & 1.000 & 1.000 & \textbf{1.000} & \textbf{0.999} & \textbf{0.998} \\
\midrule
 & Teacher      & 0.246 & 0.178 & --- & $-0.298$ & $-0.497$ & --- \\
PushT & $\lambda=0$  & 0.633 & 0.677 & --- & $-0.054$ & $-0.348$ & --- \\
 & $\lambda=16$ & \textbf{0.946} & \textbf{0.872} & --- & \textbf{0.666} & \textbf{0.139} & --- \\
\bottomrule
\end{tabular}
\end{table}

\section{Additional results}
\label{app:additional}

Material referenced from the main text.

\subsection{Symmetry breaking along data, optimization and capacity}

\begin{table}[h]
\centering
\caption{The incentive strengthens along all three axes the threshold implicates.
Cross-encoding cosine; lower is more broken.}
\label{tab:axes}
\begin{tabular}{llcc}
\toprule
Axis & Varied & From & To \\
\midrule
Data          & $n=150 \to 1200$           & \multicolumn{2}{c}{collapse $2.5\times \to 9.9\times$} \\
Optimization  & $800 \to 8000$ steps       & $\cos 0.370$ & $\cos 0.242$ \\
Capacity      & hidden $1024/2 \to 2048/3$ & $\cos 0.370$ & $\cos 0.125$ \\
\bottomrule
\end{tabular}
\end{table}

\subsection{When the orbit is too large to enumerate}
\label{app:impossible}

Eq.~\eqref{eq:repair} sums over the orbit, so it is available only when the orbit can be
enumerated. Our axis has two elements, which is why the repair is cheap; it is worth being
explicit about how quickly that stops being true, because it bounds where this procedure
applies.

\paragraph{Discrete orbits grow factorially.} Enforcing invariance to the order of $N$
interchangeable inputs requires averaging over $S_N$, which is $N! \approx 9.3\times10^{157}$
terms at $N=100$. Monte Carlo over the orbit converges as $k^{-1/2}$
(Proposition~\ref{prop:mc-accuracy}), so reducing an order-induced standard deviation of
$\sim10^{-1}$ to negligible costs on the order of $10^{14}$ forward passes for a single
prediction.

\paragraph{Continuous orbits are not finite sums at all.} A continuous group requires
integration against the Haar measure rather than a sum, and no finite sample of the orbit is
exact. A camera pose treated as a continuous nuisance is in this case; an action encoding with
two presentations is not.

\paragraph{Where that leaves robot dynamics.} The tractable cases are those where the
presentations are few and enumerable, which covers the parameterization choices a robot stack
actually makes: absolute against relative actions, one frame convention against another, a
fixed set of unit conventions. The intractable ones are those where the nuisance is continuous
or combinatorial, and for those the architectural route of Eq.~\eqref{eq:threshold},
a representation that never encodes the distinction, is the only practical answer.

\subsection{True symmetry breaking at finite $n$}

\begin{table}[htbp]
  \caption{True symmetry breaking at finite $n$. Target $Y = \mathbf{1}\{\mathrm{median}(\mathbf{x}) > 0 \;\wedge\; \mathrm{std}(\mathbf{x}) > 0.8 \;\wedge\; \mathrm{range}(\mathbf{x}) > 2\}$ is exactly $S_d$-invariant. The unconstrained model's orbit dispersion under random coordinate permutations (TV to the
  orbit mean) is large at finite $n$ and grows with $|S_d|$, confirming that compression selects non-invariant predictors even when the target is invariant.}
  \label{tab:true-symmetry}
  \centering
  \small
  \begin{tabular}{@{}rlrrrr@{}}
    \toprule
    $d$ & $|S_d|$ & $n_{\mathrm{train}}$ & Acc (unconstr.) & Acc (invariant) & Orbit disp. \\
    \midrule
    8 & $4 \times 10^4$ & 1{,}000 & 0.858 & 0.979 & 0.141 \\
    8 & $4 \times 10^4$ & 10{,}000 & 0.942 & 0.993 & 0.059 \\
    8 & $4 \times 10^4$ & 100{,}000 & 0.968 & 0.995 & 0.024 \\
    \midrule
    16 & $2.1 \times 10^{13}$ & 1{,}000 & 0.756 & 0.974 & 0.229 \\
    16 & $2.1 \times 10^{13}$ & 10{,}000 & 0.828 & 0.991 & 0.163 \\
    16 & $2.1 \times 10^{13}$ & 100{,}000 & 0.918 & 0.994 & 0.062 \\
    \midrule
    32 & $2.6 \times 10^{35}$ & 1{,}000 & 0.708 & 0.978 & 0.277 \\
    32 & $2.6 \times 10^{35}$ & 10{,}000 & 0.757 & 0.991 & 0.224 \\
    32 & $2.6 \times 10^{35}$ & 100{,}000 & 0.810 & 0.993 & 0.151 \\
    \bottomrule
  \end{tabular}
\end{table}

\subsection{Collapse ratio against training-set size}

\begin{figure}[tbp]
\centering
\includegraphics[width=\textwidth]{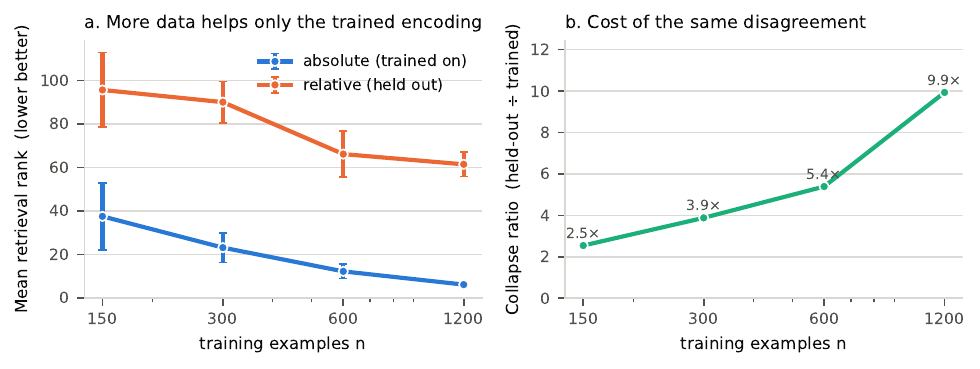}
\caption{More data improves the trained encoding and barely improves the equivalent one, so
the collapse ratio grows from $2.5\times$ to $9.9\times$. Panel b is a quotient of the two
curves in panel a, and should be read as the growing cost of a fixed disagreement, not as a
growing disagreement: cross-encoding cosine is flat across the same sweep
($0.357,0.337,0.374,0.370$).}
\label{fig:scaling}
\end{figure}

Figure~\ref{fig:scaling} gives the $n$ sweep behind \S\ref{sec:threshold}. Panel a is the
quantity to read: more data improves the encoding the teacher was trained on and leaves the
equivalent one where it was.

\subsection{Cost of orbit-constancy versus $\lambda$}

\begin{figure}[tbp]
\centering
\includegraphics[width=\textwidth]{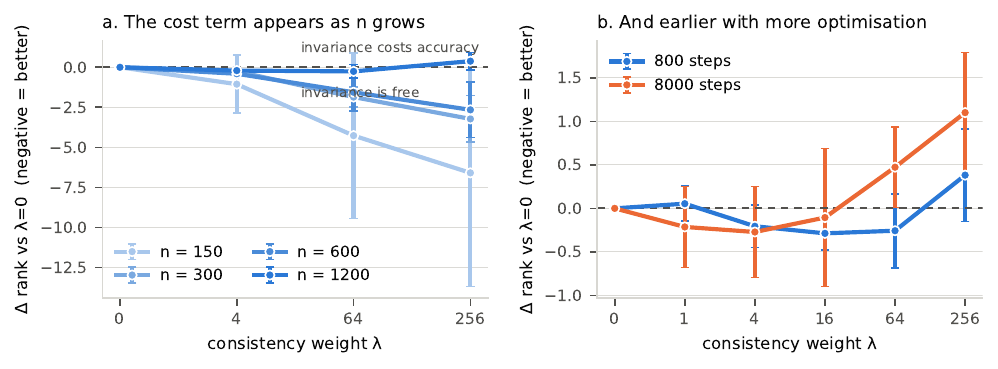}
\caption{The cost of orbit-constancy as a function of $\lambda$, relative to $\lambda=0$.
Negative is better. Left: the benefit shrinks and finally reverses as $n$ grows. Right: the
same, earlier, with more optimization.}
\label{fig:lambda}
\end{figure}

Figure~\ref{fig:lambda} sweeps the disagreement weight against training-set size, and locates
the point at which orbit-constancy stops being free regularisation and starts costing rank.

\subsection{Admissibility certificates on four numerical axes}
\label{app:certificates}

\begin{table}[tbp]
\centering
\caption{Admissibility certificates on four further numerical axes, 3 seeds each. All are
admissible; two exhibit no measurable departure because the featurization already removes the
nuisance.}
\label{tab:cross}
\begin{tabular}{llrrl}
\toprule
Axis & Family & Teacher dispersion & Reduction & Certificate \\
\midrule
Angle wrap (Kepler)        & gauge & $8.6\times10^{-15}$ & --- & bijective \\
Time origin (Lorenz)       & gauge & $4.2\times10^{-12}$ & --- & bijective \\
Circular shift             & gauge & $2.6\times10^{-2}$ & $100\%$ & quotient \\
Node permutation (Ax=b)    & exchangeability & $3.3\times10^{2}$ & $83.4\%$ & quotient \\
\bottomrule
\end{tabular}
\end{table}

\section{Description-length account of the incentive}
\label{app:formal-setup}

This appendix collects the description-length results the main text uses, in the two-part MDL
setting of \citet{grunwald} and for the strictly proper scoring rules of
\citet{gneiting}. They price the incentive to exploit a notation, and they are group-agnostic: nothing below refers to which
transformation is at stake, only that some coordinate of the input distinguishes
presentations. Proofs are omitted where they are routine.

Throughout, $G$ acts measurably on inputs, $Z$ is a maximal invariant, $\hat P$ is the
empirical law, and $\mathcal F_{\mathrm{inv}}\subseteq\mathcal F$ is the orbit-constant
subclass. The \emph{effective orbit information}
$I^{\mathcal F}_{\hat P}(Y;G\mid Z)$ of \S\ref{sec:framework} is the per-sample code length
$\mathcal F$ gives up by being restricted to $\mathcal F_{\mathrm{inv}}$, and
$\Delta_{\mathrm{mis}} := I^{\mathcal F}_{\hat P}(Y;G\mid Z)-I_{\hat P}(Y;G\mid Z)$ is its
misspecification component. Write $\mathrm{Reg}_{\hat P}(\mathcal F;X)$ for the best-in-class
regret against the Bayes envelope.

\paragraph{Three regimes, not one conclusion.} Everything downstream turns on when
$I^{\mathcal F}_{\hat P}(Y;G\mid Z)$ is positive. It is a \emph{best-in-class} gap, so by
Corollary~\ref{cor:no-sign-flip} it equals $I_{\hat P}(Y;G\mid Z)+\Delta_{\mathrm{mis}}$,
within $\varepsilon_X+\varepsilon_Z$ of the population term, and three cases follow.
(i) The group coordinate informs the target: the population term is positive and the penalty
for invariance grows linearly in $n$. This is the case of an approximate or assumed symmetry,
where the transformation is not one, and we note it only to set it aside; it is not the case
this paper studies.
(ii) The target is exactly invariant and both classes realize their infima: that term is zero,
the linear growth disappears, and what remains is $\Delta_{\mathrm{mis}}$, a finite-sample
burden growing with orbit size and decaying slowly in $n$ (Table~\ref{tab:true-symmetry}).
(iii) The labels never exercise the coordinate: the objective prices it in neither direction,
no reading of $n$ applies, and the mode is simply unconstrained. \S\ref{sec:exp} studies the
third case. The left-hand side is architectural in all three, so \textbf{a representation that
makes the invariance cheap to describe removes the incentive entirely}, which is the lever to
reach for wherever an equivariant construction exists.

\begin{definition}[Prequential (two-part) MDL]
\label{def:mdl}
Given a predictor class $\mathcal{F}$ and a description-length proxy $L(\mathcal{F})$,
\[
\MDL_n(\mathcal{F}) := L(\mathcal{F}) + n \cdot \inf_{f \in \mathcal{F}} \E_{\hat{P}}[-\log f(Y \mid X)].
\]
\end{definition}

\begin{theorem}[MDL symmetry-breaking threshold]
\label{thm:mdl-threshold}
Under Definition~\ref{def:mdl},
\[
\MDL_n(\mathcal{F}_{\mathrm{inv}}) - \MDL_n(\mathcal{F}) = \underbrace{L(\mathcal{F}_{\mathrm{inv}}) - L(\mathcal{F})}_{\text{description-length diff.}} + n \cdot I^{\mathcal{F}}_{\hat{P}}(Y; G \mid Z).
\]
Enforcing invariance is MDL-preferred iff $L(\mathcal{F}) - L(\mathcal{F}_{\mathrm{inv}}) \geq n \cdot I^{\mathcal{F}}_{\hat{P}}(Y; G \mid Z)$.
\end{theorem}

\noindent The decomposition holds for any strictly proper scoring rule, since propriety makes
functional invariance and distribution invariance coincide.

\begin{corollary}[No sign flip; Bayes-envelope approximation]
\label{cor:no-sign-flip}
Since $I^{\mathcal{F}}_{\hat{P}}(Y; G \mid Z) \geq 0$, necessarily $\Delta_{\mathrm{mis}} \geq -I_{\hat{P}}(Y; G \mid Z)$: misspecification can cancel the conditional mutual information term but cannot reverse the direction of the best-in-class incentive. If $\mathrm{Reg}_{\hat{P}}(\mathcal{F}; X) \leq \varepsilon_X$ and $\mathrm{Reg}_{\hat{P}}(\mathcal{F}_{\mathrm{inv}}; Z) \leq \varepsilon_Z$, then $|I^{\mathcal{F}}_{\hat{P}}(Y; G \mid Z) - I_{\hat{P}}(Y; G \mid Z)| \leq \varepsilon_X + \varepsilon_Z$.
\end{corollary}

\noindent This is what makes the three regimes of \S\ref{sec:framework} exhaustive: the
best-in-class gap is the population conditional mutual information plus a misspecification
term that is bounded but need not vanish.

\subsection{Orbit-constancy}
\label{app:orbit-constancy}

\begin{theorem}[Optimal prediction is constant on symmetry orbits]
\label{thm:orbit-constancy}
If $p^*(y \mid x)$ is $G$-invariant and $f^*$ is the Bayes-optimal predictor under any strictly proper scoring rule, then $f^*(y \mid x) = f^*(y \mid g \cdot x)$ for all $g \in G$, for all $x, y$.
\end{theorem}

\noindent This is the licence for treating orbit dispersion as an error measure at all: under
a genuine symmetry, any dispersion is a departure from the optimum. It is also why
certification (\S\ref{sec:admissible}) is load bearing, since the theorem says nothing when
$p^*$ is not $G$-invariant.

\begin{proposition}[Monte Carlo accuracy of orbit averaging]
\label{prop:mc-accuracy}
Define $U := f(y \mid g \cdot x)$ for $g \sim \mu$ and $\bar{U}_K := \frac{1}{K}\sum_{k=1}^K f(y \mid g_k \cdot x)$. If $\mathrm{Var}(U) < \infty$ and $g_1, \ldots, g_K \overset{\mathrm{i.i.d.}}{\sim} \mu$, then $\mathrm{Var}(\bar{U}_K) = \mathrm{Var}(U)/K$ and $\mathrm{Std}(\bar{U}_K) = \mathrm{Std}(U)/\sqrt{K}$.
\end{proposition}

\section{Proof of Theorem~\ref{thm:parity}}
\label{app:parity}

Let $\varphi_1,\dots,\varphi_K:\mathcal X\to\mathcal V$ be the view family and let $\sim$ be
the equivalence relation on $\mathcal V$ generated by $\varphi_v(x)\sim\varphi_w(x)$ for all
$x\in\mathcal X$ and all $v,w$. Write $q:\mathcal V\to\mathcal V/\!\sim$ for the quotient map
and $Q:=q(\varphi_1(X))$; by construction of $\sim$ we have $q(\varphi_v(X))=Q$ for every $v$,
so $Q$ is well defined without reference to which presentation was received.

\paragraph{Step 1: orbit-constant predictors are exactly the $Q$-measurable ones.} A predictor
$f$ on $\mathcal V$ satisfies $f(\varphi_v(x))=f(\varphi_w(x))$ for all $x,v,w$ if and only if
$f$ is constant on $\sim$-classes, which is to say $f=\tilde f\circ q$ for some $\tilde f$.
Minimising log loss over this class gives the conditional law of $Y$ given $Q$, so
\[
R_{\mathrm{inv}} \;=\; \inf_{\tilde f}\,\E\big[-\log \tilde f(Y\mid Q)\big] \;=\; H(Y\mid Q).
\]
Because $f$ is orbit-constant the value does not depend on which $v$ generates the input, so
no mixing distribution over $v$ need be specified.

\paragraph{Step 2: $Q$ is a coarsening of every view.} For each $v$, $Q=q(\varphi_v(X))$ is a
measurable function of $\varphi_v(X)$, hence
$\sigma(Q)\subseteq\sigma(\varphi_v(X))$. Conditioning on a coarser $\sigma$-algebra cannot
decrease conditional entropy, so
\[
R_{\mathrm{inv}} \;=\; H(Y\mid Q)\;\ge\;H(Y\mid \varphi_v(X))\;=\;R(\varphi_v)
\qquad\text{for every } v .
\]

\paragraph{Step 3.} Taking the maximum over $v$ gives
$R_{\mathrm{inv}}\ge\max_v R(\varphi_v)$, and subtracting $\min_v R(\varphi_v)$ from both
sides gives $R_{\mathrm{inv}}-\min_v R(\varphi_v)\ge\delta$. \hfill$\square$

\paragraph{Equality, and the two certificates.} If the views are mutually reconstructible,
each $\varphi_v(x)$ determines every $\varphi_w(x)$, so each $\sim$-class meets each view's
range once and $\sigma(Q)=\sigma(\varphi_v(X))$. Step 2 is then an equality and
$R_{\mathrm{inv}}=R(\varphi_v)$: a bijective relabelling costs nothing. Reconstructibility is
not needed for that conclusion. If $\varphi_w=c\circ\varphi_v$ for a coarsening $c$ that is
sufficient for $Y$, meaning $H(Y\mid\varphi_v(X))=H(Y\mid c(\varphi_v(X)))$, then $\delta=0$
and the bound is again tight at zero even though $c$ is not invertible. Angle wrap is this
case, which is why the test reports reconstructibility as a certificate and gates only on
parity.

\paragraph{Beyond log loss.} Only two properties of log loss were used: that its Bayes risk is
the conditional entropy, and that this risk is concave in the predictive distribution so
coarsening cannot decrease it. Both hold for the generalized entropy of any strictly proper
scoring rule, so the statement carries over with $H$ replaced by that rule's Bayes risk.

\section{Direction-valued prediction has no Jensen guarantee}
\label{app:diagnostics}

Fix an input $x$ and nuisance sampler $\mu$. Given $K$ samples $g_1,\ldots,g_K \sim \mu$, let $\bar{p} := \frac{1}{K}\sum_{k=1}^K f(\cdot \mid g_k \cdot x)$. The empirical Jensen gain
\[
\hat{J}(x; y) := \frac{1}{K}\sum_{k=1}^K \big(-\log f(y \mid g_k \cdot x)\big) - \big(-\log \bar{p}(y)\big) \geq 0
\]
is non-negative by concavity of the logarithm. This is why orbit
averaging is safe wherever the prediction is a probability, and it is the guarantee the
setting is not one where that guarantee applies.

The guarantee needs the prediction to be a probability vector and the score to be log loss.
Neither holds when the prediction is a unit vector scored by cosine similarity, and
Proposition~\ref{prop:no-jensen} is the counterexample. Take $d=2$ and $y=(1,0)$, with orbit
predictions
\[
u_1=(1,0),\qquad u_{2,3}=\big(-\tfrac{9}{10},\ \pm\sqrt{1-\tfrac{81}{100}}\big),
\]
all of unit norm. The individual scores are $\langle u_1,y\rangle = 1$ and
$\langle u_{2},y\rangle=\langle u_{3},y\rangle=-\tfrac{9}{10}$. Their sum is
$\sum_k u_k = (-\tfrac{8}{10},0)$, so $\bar u = (-1,0)$ and $\langle\bar u,y\rangle=-1$,
strictly below all three. The inequality $-1 < -\tfrac{9}{10}$ is strict, so the witness
survives any sufficiently small perturbation of $y$ and of the $u_k$ and is not an artifact of
exact symmetry. Taking instead $u_{1,2}=(0,\pm1)$ gives $\sum_k u_k = 0$, where $\bar u$ is
undefined; that configuration is not itself stable, but $\|\sum_k u_k\|$ is arbitrarily small
in a neighbourhood of it, so the normalized mean is ill-conditioned there. Both failures are
properties of the output space rather than of any particular predictor. They rule out
averaging the outputs, not averaging as such, which is why Eq.~\eqref{eq:repair} keeps the
average and moves it to the loss.

\paragraph{Audit and repair.} Orbit averaging supplies a natural teacher, and a student can be
distilled to match it. We use that as the falsifiable make/break test of
Definition~\ref{def:makebreak}: if distillation reduces dispersion without materially harming
task performance, the nuisance family is a plausible invariance constraint, and otherwise it
is rejected. Algorithm~\ref{alg:audit-repair} states the procedure; for direction-valued
predictions the distillation step is replaced by Eq.~\eqref{eq:repair} for the reason above.

\begin{algorithm}[t]
\caption{Audit and repair for a certified symmetry}
\label{alg:audit-repair}
\begin{algorithmic}[1]
\REQUIRE base model $f_T$, nuisance sampler $\mu$, orbit samples $K$, metric $M$, dispersion statistic $A$, tolerance $\varepsilon$
\STATE Certify the transformation family (\S\ref{sec:admissible}); abort unless predictive parity holds
\STATE Measure $A(f_T)$ and $M(f_T)$ on held-out data
\STATE Fit $g_\theta$ by Eq.~\eqref{eq:repair} for probability- or direction-valued outputs
\STATE Bootstrap confidence intervals for $\Delta M$ and $\Delta A$
\RETURN \textsc{make} if $\mathrm{CI}[\Delta A] < 0$ and $\mathrm{CI}[\Delta M] \geq -\varepsilon$, else \textsc{break}
\end{algorithmic}
\end{algorithm}

\section{Supplementary experiments}
\label{app:supplementary}

\subsection{Action-conditioned dynamics: robotics and iPhyRE}
\label{app:robotics}

Three robotics tasks (ALOHA, LIBERO-10, PushT) and one physics simulation (iPhyRE). Frozen ViT-H/14 JEPA encoder; predictor receives $(z_t, a_t)$ and predicts future latent. Evaluation by retrieval (lower mean rank is better). Key intervention: action-shuffle (deliberately not evidence-preserving).

\begin{table}[htbp]
  \caption{Action conditioning: action-shuffle degrades performance, confirming causal use.}
  \label{tab:action}
  \centering
  \small
  \begin{tabular}{@{}lcccccc@{}}
    \toprule
    Task & rank$\downarrow$ (no-act) & rank$\downarrow$ (act) & rank$\downarrow$ (shuf) & R@10$\uparrow$ (no-act) & R@10$\uparrow$ (act) & Verdict \\
    \midrule
    ALOHA     & 186.2 & 97.7  & 635.8 & 0.068 & 0.124 & \textsc{make} \\
    LIBERO-10 & 80.9  & 49.6  & 55.5  & 0.136 & 0.242 & \textsc{make} \\
    PushT     & 179.9 & 349.9 & 547.7 & 0.079 & 0.093 & \textsc{break} \\
    \bottomrule
  \end{tabular}
\end{table}

Action conditioning produces large improvements on ALOHA and LIBERO-10; action-shuffle degrades performance below baseline, confirming causal use of the action signal. PushT yields \textsc{break}: adding actions hurts retrieval rank ($179.9 \to 349.9$), indicating the action representation is incompatible with the encoder's latent geometry. iPhyRE shows the clearest separation: action conditioning improves mean rank $5.5\times$ ($172.6 \to 31.3$); shuffled actions revert to baseline ($173.9$).

\begin{table}[htbp]
  \caption{iPhyRE: $5.5\times$ retrieval improvement with action conditioning.}
  \label{tab:iphyre}
  \centering
  \small
  \begin{tabular}{@{}lcccc@{}}
    \toprule
    Metric & persist & no action & action & shuf \\
    \midrule
    mean rank$\downarrow$ & 167.5 & 172.6 & 31.3 & 173.9 \\
    Recall@10$\uparrow$   & 0.203 & 0.185 & 0.471 & 0.124 \\
    \bottomrule
  \end{tabular}
\end{table}

\subsection{Molecular dynamics: representation ablation}
\label{app:molecular}

Lennard-Jones fluid with exact SE(3)${\times}S_n$ symmetry. Two encoders: RDF (invariant) vs.\ flattened coordinates (not aligned). Same data, same training, same transformation family ($K = 16$).

The RDF encoder is SE(3)-invariant by construction: orbit dispersion is already small (0.012--0.017) and distillation preserves accuracy ($0.894 \to 0.893$). The flattened-coordinate encoder breaks SE(3); orbit-averaging over rotations and permutations reduces accuracy catastrophically ($0.856 \to 0.456$) because the representation is not aligned with the symmetry. Same physical system, same symmetry group, same training. The representation determines the verdict.

\begin{table}[htbp]
  \caption{Same symmetry, two representations. The defect is caused by the representation.}
  \label{tab:molecular}
  \centering
  \small
  \begin{tabular}{@{}lccccc@{}}
    \toprule
    Encoder & Acc (base) & Acc (stud.) & TV-mean (base) & TV-mean (stud.) & Verdict \\
    \midrule
    RDF ($\sigma{=}0.01$)     & 0.894 & 0.893 & 0.017 & 0.016 & \textsc{make} \\
    RDF ($\sigma{=}0.03$)     & 0.894 & 0.893 & 0.012 & 0.011 & \textsc{make} \\
    Flatten ($\sigma{=}0.01$) & 0.856 & 0.456 & 0.449 & 0.433 & \textsc{break} \\
    \bottomrule
  \end{tabular}
\end{table}


\begin{thebibliography}{99}

\bibitem[Arjovsky et al.(2019)]{irm}
M.~Arjovsky, L.~Bottou, I.~Gulrajani, and D.~Lopez-Paz.
\newblock Invariant risk minimization, 2019.
\newblock arXiv:1907.02893.

\bibitem[Peters et al.(2016)]{peters2016}
J.~Peters, P.~B\"uhlmann, and N.~Meinshausen.
\newblock Causal inference by using invariant prediction: identification and confidence intervals.
\newblock \emph{J. R. Stat. Soc. B}, 78(5):947--1012, 2016.


\bibitem[Assran et al.(2023)]{ijepa}
M.~Assran, Q.~Duval, I.~Misra, P.~Bojanowski, P.~Vincent, M.~Rabbat, Y.~LeCun, and N.~Ballas.
\newblock Self-supervised learning from images with a joint-embedding predictive architecture.
\newblock In \emph{CVPR}, 2023.

\bibitem[Assran et al.(2025)]{vjepa2}
M.~Assran et al.
\newblock V-JEPA 2: self-supervised video models enable understanding, prediction and planning, 2025.
\newblock arXiv:2506.09985.

\bibitem[Cadene et al.(2024)]{lerobot}
R.~Cadene, S.~Alibert, A.~Soare, Q.~Gallouedec, A.~Zouitine, and T.~Wolf.
\newblock LeRobot: state-of-the-art machine learning for real-world robotics in PyTorch, 2024.
\newblock \url{https://github.com/huggingface/lerobot}.

\bibitem[Hafner et al.(2023)]{dreamerv3}
D.~Hafner, J.~Pasukonis, J.~Ba, and T.~Lillicrap.
\newblock Mastering diverse domains through world models, 2023.

\bibitem[Hansen et al.(2024)]{tdmpc2}
N.~Hansen, H.~Su, and X.~Wang.
\newblock TD-MPC2: scalable, robust world models for continuous control.
\newblock In \emph{ICLR}, 2024.

\bibitem[LeCun(2022)]{lecun2022}
Y.~LeCun.
\newblock A path towards autonomous machine intelligence, 2022.

\bibitem[Zhao et al.(2023)]{aloha}
T.~Z.~Zhao, V.~Kumar, S.~Levine, and C.~Finn.
\newblock Learning fine-grained bimanual manipulation with low-cost hardware.
\newblock In \emph{RSS}, 2023.

\bibitem[Chi et al.(2023)]{pusht}
C.~Chi, S.~Feng, Y.~Du, Z.~Xu, E.~Cousineau, B.~Burchfiel, and S.~Song.
\newblock Diffusion policy: visuomotor policy learning via action diffusion.
\newblock In \emph{RSS}, 2023.

\bibitem[Cohen and Welling(2016)]{cohen2016}
T.~S.~Cohen and M.~Welling.
\newblock Group equivariant convolutional networks.
\newblock In \emph{ICML}, 2016.

\bibitem[Satorras et al.(2021)]{egnn}
V.~G.~Satorras, E.~Hoogeboom, and M.~Welling.
\newblock E(n) equivariant graph neural networks.
\newblock In \emph{ICML}, 2021.

\bibitem[Elesedy and Zaidi(2021)]{elesedy2021}
B.~Elesedy and S.~Zaidi.
\newblock Provably strict generalisation benefit for equivariant models.
\newblock In \emph{ICML}, 2021.

\bibitem[Chen et al.(2020)]{chen2020}
S.~Chen, E.~Dobriban, and J.~H.~Lee.
\newblock A group-theoretic framework for data augmentation.
\newblock In \emph{NeurIPS}, 2020.

\bibitem[Grünwald(2007)]{grunwald}
P.~Grünwald.
\newblock \emph{The Minimum Description Length Principle}.
\newblock MIT Press, 2007.

\bibitem[Gneiting and Raftery(2007)]{gneiting}
T.~Gneiting and A.~E.~Raftery.
\newblock Strictly proper scoring rules, prediction, and estimation.
\newblock \emph{JASA}, 102(477):359--378, 2007.

\end{thebibliography}
\end{document}